\documentclass[letterpaper]{article}
\usepackage{iccc}
\usepackage{url}
\usepackage{times}
\usepackage{helvet}
\usepackage{courier}
\usepackage{amssymb} 
\usepackage{graphicx}
\usepackage{booktabs}
\usepackage{amsmath}
\usepackage{float}
\usepackage{multirow}
\usepackage{url}
\title{Anchor-Conditioned Compositional Control for
Landscape Image Generation%
}
\author{Gadha Lekshmi P,$^1$ Govind Arun,$^2$ Rohith Syam,$^3$ Ahmed Elgammal$^1$\\
$^1$Rutgers University--New Brunswick, USA \quad
$^2$University of Maryland--College Park, USA \quad
$^3$University of Technology Sydney, Australia\\
gadhalekshmip@gmail.com, govind123.ga@gmail.com,
rohithlayana@gmail.com, elgammal@cs.rutgers.edu\\
}
\begin{document} 
\maketitle
\begin{abstract}
Image generative models, though widely used as creative tools, offer limited support for the kind of compositional control that photographers and visual artists routinely exercise. This paper presents early results on an anchor-conditioned finetuning framework for landscape image generation, in which a four dimensional compositional anchor vector is extracted from training images and injected into a diffusion model via a decoupled cross-attention mechanism with Fourier encoding and three-way classifier-free guidance dropout. Quantitative evaluation against a baseline and three ablation variants shows that the proposed architecture achieves the highest horizon detection rate (0.850) and the highest rule-of-thirds alignment (0.817). A category specific ablation further demonstrates that training on compositionally homogeneous scene subsets reduces horizon deviation by up to 40\% compared to mixed training. This establishes that compositional control precision is category-dependent.
\end{abstract}

\section{Introduction}

Image generative models, while introduced as creative tools, lack controllability over composition, a capability that is fundamental to the creative process. A photographer composing a landscape does not just specify content. They decide where
the horizon sits, how the frame is divided, and where the viewer's eye should land. Diffusion models \cite{ho2020,rombach2022} make none of these decisions explicitly. They emerge from statistical patterns absorbed during pretraining, giving the photographer no reliable handle on the spatial
outcome.

The goal of this paper is to introduce a mechanism by which compositional intent can be specified, injected, and measured in a finetuned diffusion model for landscape image generation. A four dimensional anchor vector encoding horizon position, detection confidence, average saliency, and foreground ratio is automatically extracted from each training image using Hough-transform horizon detection and spectral residual saliency analysis,
requiring no manual annotation. This vector is injected into a finetuned diffusion model through a dedicated attention pathway, so that a target horizon position and compositional weighting can be specified at inference time without manual spatial masks or structural conditioning inputs.

The research demonstrates early results on this mechanism, evaluated using horizon placement accuracy and rule-of-thirds alignment as measurable proxies for compositional control. A category specific ablation investigates whether training on compositionally homogeneous scene subsets improves conditioning precision beyond what mixed training achieves.

The remainder of this paper covers related work on spatial
conditioning in diffusion models, followed by the
methodology and experimental setup, quantitative results
across four model configurations and three landscape
categories, and a discussion of findings and planned
extensions.
\section{Background}

Prior work has approached spatial control through structural conditioning inputs. ControlNet \cite{zhang2023} introduces a parallel trainable encoder that accepts depth maps and edge maps, demonstrating that a frozen pretrained model can accommodate new spatial conditioning without losing generative quality. GLIGEN \cite{li2023} extends this to open-set grounded generation using bounding boxes and keypoints, and provides the strongest empirical support for Fourier encoding of spatial coordinates, showing 6.8 times better localisation control compared to raw MLP encoding of the same values. Approaches such as SDXL \cite{podell2024} address
high-resolution generation through architectural scaling
and resolution conditioning, but do not provide mechanisms
for specifying or preserving compositional intent across
pipeline stages. IP-Adapter \cite{ye2023} introduced the decoupled cross-attention pattern that this architecture directly adopts, where the conditioning signal has its own Key and Value projections rather than competing with text tokens for the available attention budget. This work applies these ideas to the specific problem of compositional intent in landscape image generation, extending the decoupled adapter pattern from image-based conditioning to compact scalar compositional descriptors. Recent work such as Bokeh Diffusion \shortcite{atfortes2025} has similarly shown that scalar photographic parameters can be injected through lightweight cross-attention adapters, though none of these efforts address compositional layout descriptors or their preservation across generation stages.

\section{Methodology \& Experimental Setup}
\subsection{Dataset}

Training was conducted using the Cropped-1901 Landscape Dataset \cite{dxtinction2023}, which consists of 4,713 images curated from multiple publicly available sources, including a Flickr image dataset \cite{hsankesara_flickr}, a landscape image collection from Kaggle \cite{arnaud_landscape}, the Nature dataset hosted on Hugging Face \cite{mertcobanov_nature}, and the HQ-50K dataset \cite{yangqiee_hq50k}. The dataset was standardised to a 1.90:1 aspect ratio at a resolution of 1949×1024 pixels, and for training, images were resized to 768×400 pixels to maintain compatibility with the base model’s latent representation. Existing captions were wrapped with a photorealistic prefix and suffix to anchor the model to the photographic domain. The dataset was split using a 90:10 ratio, resulting in 4,241 training images and 472 validation images.
\subsection{Compositional Anchor Vector}

Before the training phase, each image in the dataset is analysed to produce a four dimensional anchor vector representing its compositional structure. The horizon position (horizon\_y) is estimated using a Hough line transform applied to a horizontal-line-biased Canny edge map of the upper 60\% of the image, normalised to $[0, 1]$ where 0 represents the top edge. The horizon confidence (horizon\_conf) is derived from the fractional coverage of the longest detected horizontal line relative to image width. The average saliency (avg\_saliency) is computed via spectral residual analysis across the top detected salient regions. The foreground ratio (fg\_ratio) provides a coarse estimate of visually prominent foreground content. When no horizon is detected, horizon\_y defaults to 0.5 and horizon\_conf to 0.0, preserving the informativeness of the remaining two components. Anchor vectors are pre-extracted for the entire dataset in a single offline pass and cached to disk.

\subsection{Pipeline}

\begin{figure}[h]
\centering
\includegraphics[width=1.1\columnwidth]{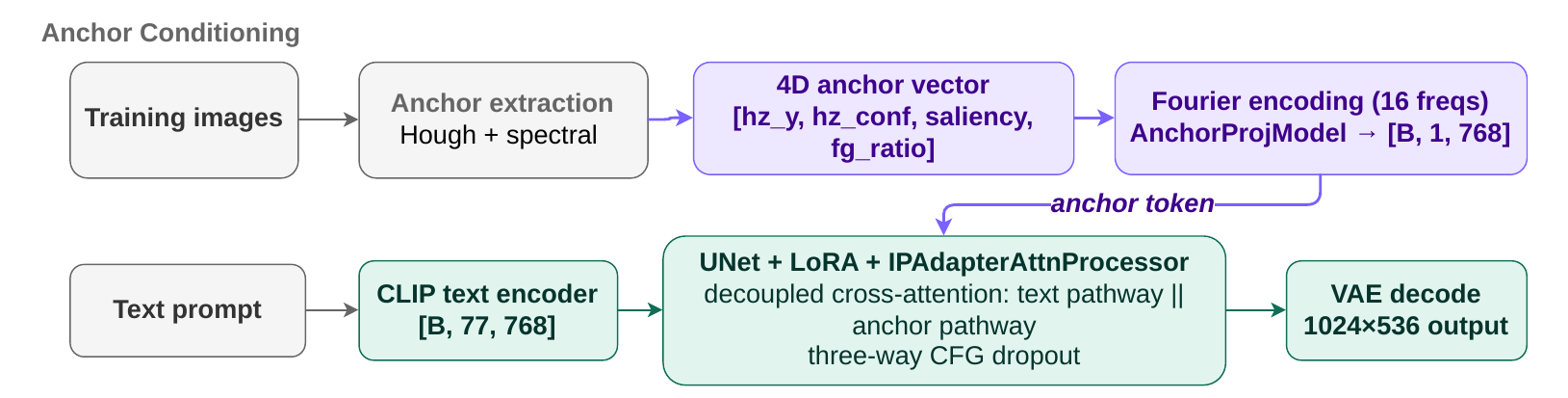}
\caption{Anchor-conditioned generation pipeline.}
\label{fig:pipeline}
\end{figure}
Figure~\ref{fig:pipeline} shows the anchor conditioned generation pipeline. The extracted anchor vector is injected into the diffusion model through a decoupled cross attention pathway.

The architecture injects compositional anchor information into a fine tuned latent diffusion model \cite{rombach2022} through a dedicated parallel attention pathway that operates independently of the text conditioning stream. This design is motivated by the finding that injecting an anchor token directly into the text sequence gives it insufficient influence to affect spatial composition: a single token representing 1.3\% of the total attention budget cannot compete with 77 deeply pretrained text tokens, as confirmed by the concatenation ablation in Section Results.

\textbf{Fourier Encoding:} The 4D anchor vector is first passed through a Fourier embedding layer applying 16 log-spaced frequencies from $2^0$ to $2^{15}$ via sine and cosine transformations, producing a 132-dimensional representation. This encoding enables learning of fine-grained spatial relationships from scalar inputs, motivated directly by the GLIGEN ablation result \cite{li2023}.

\textbf{AnchorProjModel:} The 132-dimensional Fourier representation is projected through a network consisting of a Linear layer (132 to 256), GELU activation, LayerNorm, a second Linear layer (256 to 768), and a final LayerNorm, producing a single anchor token of shape $[B, 1, 768]$. This network contributes approximately 233,000 parameters and operates in float32 precision throughout.

\textbf{Decoupled Cross-Attention:} Separate Key and Value projection matrices are introduced at each of the 16 cross-attention layers in the UNet using a low-rank decomposition of rank 8, totalling approximately 396,000 parameters. The output at each layer is:

\begin{equation}
\text{out} = \text{Attn}(Q, K_{\text{text}}, V_{\text{text}})
+ \lambda \cdot \text{Attn}(Q, K_{\text{anchor}},
V_{\text{anchor}})
\end{equation}

where $Q$ is shared between both pathways. The up-projection matrices are initialised from the pretrained UNet's text K/V weights via SVD decomposition, initialised to match the existing cross-attention subspace. At inference, three UNet forward passes are computed per denoising step:

\begin{equation}
\hat{\epsilon} = \epsilon_\varnothing
+ \lambda_{\text{text}}(\epsilon_{\text{text}} -
\epsilon_\varnothing)
+ \lambda_{\text{anchor}}(\epsilon_{\text{full}} -
\epsilon_{\text{text}})
\end{equation}

\textbf{Training:} During training, a three-way dropout scheme is applied following
classifier-free guidance \cite{ho2022cfg}: with 5\% probability both text and anchor are dropped, with 5\% probability only text is dropped, with 5\% probability only the anchor is dropped, and with 85\% probability both are retained. The model was trained on landscape images at 768x400 resolution for 10,000 steps with a learning rate of $1 \times 10^{-4}$, cosine decay with 500 warmup steps, AdamW optimisation with weight decay 0.01, and gradient clipping at norm 1.0. LoRA adapters of rank 16 with alpha 16 are applied to the four attention projection matrices. Mixed precision was BF16.

\footnotetext{Code available at: \url{https://github.com/gadhalekshmip/Anchor-Conditioned-Compositional-Control-}}
\subsection{Implementation Details}

Experiments were conducted on Google Colab using an NVIDIA A100 GPU with 40 GB of VRAM for training, ablation experiments, and evaluation. The software stack
used PyTorch 2.x, Hugging Face Diffusers, PEFT for LoRA
management, and safetensors for weight serialisation. Anchor
extraction used OpenCV's Hough line transform for horizon
detection and spectral residual saliency for salient region
identification. OpenCLIP ViT-B/32 \cite{cherti2022reproducible}was used for CLIP score
computation. Table~\ref{tab:anchorproj} details the
AnchorProjModel architecture.

\begin{table}[H]
\centering
\caption{AnchorProjModel architecture. Input is the
132-dimensional Fourier-encoded anchor; output is a single
token of shape $[B, 1, 768]$ injected into the decoupled
cross-attention pathway.}
\label{tab:anchorproj}
\small
\begin{tabular}{lcc}
\toprule
\textbf{Layer} & \textbf{In} & \textbf{Out} \\
\midrule
FourierEmbedding (16 freqs) & 4   & 132 \\
Linear                      & 132 & 256 \\
GELU + LayerNorm            & 256 & 256 \\
Linear                      & 256 & 768 \\
LayerNorm                   & 768 & 768 \\
\midrule
Total parameters & \multicolumn{2}{c}{233,088} \\
\bottomrule
\end{tabular}
\end{table}

\section{Results}
\label{sec:result}
\subsection{Training}

Validation was performed every 500 steps using four fixed
prompts with a target anchor of [0.333, 0.9, 0.3, 0.15].
Horizon deviation from the rule-of-thirds target (0.333)
was measured at each checkpoint. Table~\ref{tab:training} reports validation metrics across checkpoints

\begin{table}[h]
\centering
\caption{Validation metrics at each training checkpoint.
Detection rate is the fraction of validation prompts where
a horizon was successfully extracted from the generated
image.}
\label{tab:training}
\small
\begin{tabular}{ccccc}
\toprule
\textbf{Step} & \textbf{Det.} & \textbf{Mean Hz} &
\textbf{Std} & \textbf{Mean Dev.} \\
\midrule
500    & 100\% & 0.465 & 0.148 & 0.191 \\
2,000  & 75\%  & 0.302 & 0.002 & 0.031 \\
5,000  & 100\% & 0.341 & 0.091 & 0.074 \\
7,000  & 100\% & 0.305 & 0.008 & 0.028 \\
10,000 & 100\% & 0.300 & 0.003 & 0.033 \\
\bottomrule
\end{tabular}
\end{table}

\begin{figure}[h]
\centering
\includegraphics[width=\columnwidth]{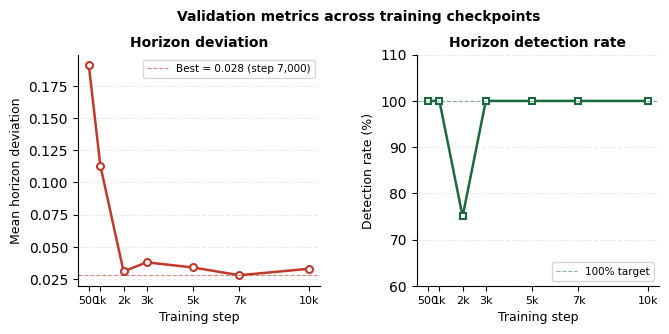}
\caption{Validation horizon deviation (left) and detection
rate (right) across training steps. Deviation improves
from 0.191 at step 500 to 0.028 at step 7,000 and then
stabilises}
\label{fig:traincurve}
\end{figure}

Mean deviation improved from 0.191 at step 500 to 0.028
at step 7,000, representing approximately a 7-fold
improvement in compositional precision. The model
stabilised around a horizon position of 0.300 from step
3,000 onwards, confirming that
the anchor conditioning converges cleanly. Figure
~\ref{fig:traincurve} shows the deviation curve across
training. The transient drop in
detection rate at step 2,000 reflects a mid-training
transition phase and does not persist.

\subsection{Compositional Control Evaluation}

Four model configurations were evaluated on 10 prompts across 6 landscape categories (mountain, ocean, forest, desert, arctic, canyon) with 6 random seeds per prompt, yielding 60 images per model at 1024x536 resolution. The configurations are: an unmodified baseline with no finetuning; a LoRA-only finetune with no anchor conditioning; a concatenation-based anchor variant appending the anchor token directly to the 77-token text sequence; and the proposed decoupled cross attention architecture. Here, horizon deviation is measured across diverse prompt categories rather than the four fixed validation prompts used during training, which accounts for the difference between training-time deviation and the values reported below. 
\begin{table}[h]
\centering
\caption{Four-model comparative evaluation. All metrics are averaged over 60 images per model.}
\label{tab:fourmodel}
\small
\begin{tabular}{lcccc}
\toprule
\textbf{Metric} & \textbf{Baseline} & \textbf{LoRA} &
\textbf{Concat} & \textbf{Proposed} \\
\midrule
Detection Rate ($\uparrow$)  & 0.766 & 0.816 & 0.766 & \textbf{0.850} \\
Horizon Dev. ($\downarrow$)  & 0.073 & 0.077 & 0.073 & \textbf{0.062} \\
Rule-of-Thirds ($\uparrow$)  & 0.500 & 0.636 & 0.500 & \textbf{0.817} \\
Sharpness ($\uparrow$)       & 2171  & 1078   & 2171  & \textbf{4606}  \\
CLIP Score ($\uparrow$)      & 0.294 & 0.280 & 0.294 & \textbf{0.297} \\
NIQE ($\downarrow$)          & 8.79  & 8.51  & 8.79  & \textbf{7.44}  \\

\bottomrule
\end{tabular}
\end{table}
\begin{figure}[H]
\centering
\includegraphics[width=\columnwidth]{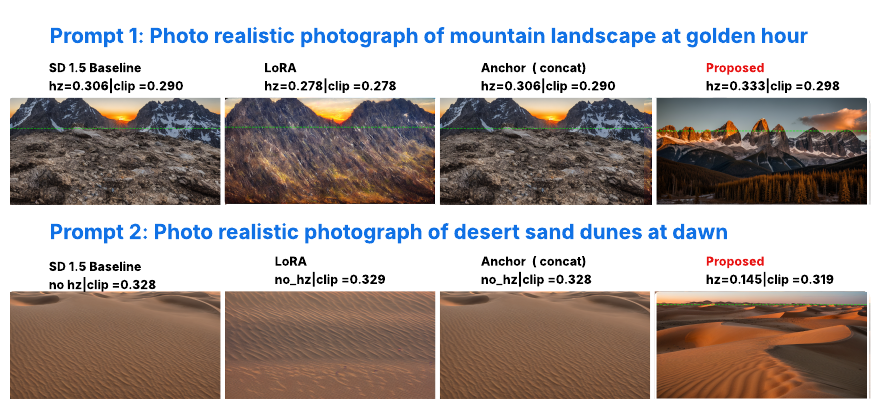}
\caption{Four-model visual comparison for two representative prompts. \textit{Top row:} mountain landscape, generated with target horizon\_y = 0.333. The proposed model achieves exact rule-of-thirds placement \textit{Bottom row:} desert sand dunes, where rest of the models fail to produce a detectable horizon; the proposed model successfully places the horizon, consistent with a high-sky compositional target.}
\label{fig:fourmodel}
\end{figure}

The proposed architecture achieves the best performance across all primary compositional metrics. Table~\ref{tab:fourmodel} and Figure~\ref{fig:fourmodel} show the full comparison. The concatenation variant scores
identically to the unmodified baseline on every metric, confirming that a single appended anchor token is completely ignored by the model: it represents only 1.3\% of the total attention budget and cannot influence spatial
layout when competing with 77 pretrained text tokens. The LoRA-only model moderately improves horizon detection, reflecting that domain finetuning absorbs some implicit
compositional patterns, but its horizon deviation remains
marginally worse than baseline, indicating these patterns
do not generalise to precise compositional control. The
proposed architecture's horizon deviation of 0.062
represents a 15\% reduction in compositional error
relative to baseline, and its rule-of-thirds alignment
of 0.817 is 63\% higher, indicating substantially more
frequent placement of salient regions near compositional power points.

\subsection{Category-Specific Ablation}

\begin{figure}[H]
\centering
\includegraphics[width=\columnwidth]{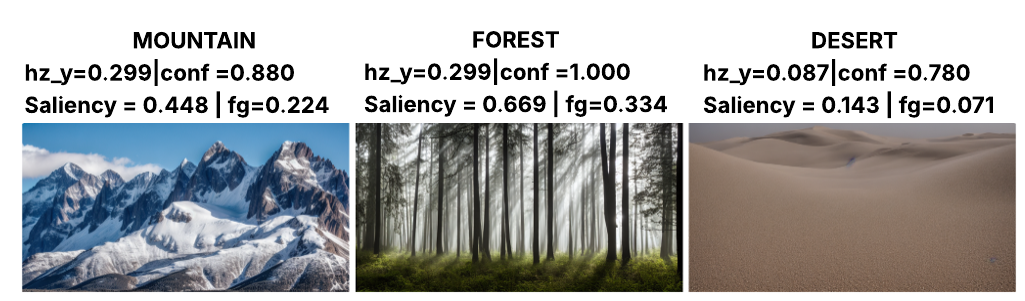}
\caption{Sample outputs from the three category specific models (mountain, forest, desert), each annotated with the anchor values extracted from the generated image.}
\label{fig:categorysamples}
\end{figure}

To investigate whether compositional control is category-dependent, the 4,713 training images were classified into 3 scene categories using keyword matching on existing dataset captions followed by CLIP zero-shot verification for ambiguous cases. Three categories: mountain (1,534 images), forest (1,149 images), and desert (1,042 images). Each category specific model used the identical architecture and training configuration as the proposed model, with step budget scaled proportionally to subset size (1,301 steps). Each model was evaluated on 100 images generated from 10 category-matched prompts with 10 random seeds, ensuring the evaluation prompts align with the training domain of each model. Table~\ref{tab:ablation} and Figure~\ref{fig:categorysamples} report the per-category results.

\begin{table}[h]
\centering
\caption{ All
category-specific models share identical architecture and
hyperparameters with the proposed model. Best step is selected by lowest validation horizon deviation.}
\label{tab:ablation}
\small
\begin{tabular}{lcccc}
\toprule
\textbf{Model} & \textbf{Images} & \textbf{Steps} &
\textbf{Det. ($\uparrow$)} & \textbf{Hz Dev. ($\downarrow$)} \\
\midrule
Proposed (mixed) & 4,713 & 10,000 & 0.833 & 0.048 \\
Mountain         & 1,534 & 1,301 & 1.000 & 0.033 \\
Forest           & 1,149 & 1,301 & 1.000 & 0.029\\
Desert           & 1,042 & 1,301 & 1.000 & 0.033 \\
\bottomrule
\end{tabular}
\end{table}

All three category specific models achieve 100\% horizon detection and horizon deviation substantially below both the unmodified baseline and the mixed training model. The forest-specific model achieves the lowest deviation of 0.029, representing a 60\% improvement over baseline and a 40\%
improvement over the proposed mixed model. Desert training produces the most stable convergence, with deviation constant at 0.033 across both checkpoints at step 500 and step 1,000, indicating the model has cleanly learned the desert compositional prior without overfitting. The consistent improvement across all three categories, each trained with identical architecture and hyperparameters, confirms that category specific training reliably improves anchor conditioning precision.

\section{Discussion and Conclusion}

The four model evaluation shows that decoupled cross
attention is necessary for effective anchor conditioning. The concatenation ablation, which scores identically to the unmodified baseline on every metric, establishes that adding an
anchor token to the text sequence produces no measurable
effect. The proposed architecture resolves this by giving
the anchor its own Key and Value projections \cite{ye2023},
conditioning layout through a separate pathway with no
competition for attention capacity.

The category-specific ablation reveals a finding with
practical implications for finetuning strategy. Training
on a compositionally homogeneous subset consistently
outperforms training on the full mixed dataset with the
same architecture and fewer steps, confirming that
category-specific training reliably tightens conditioning
precision when the intended application domain is known
in advance.

\textbf{Future work.} These are early results on an
approach we intend to develop further. The current
evaluation relies on automatic metrics that proxy
compositional quality through horizon placement and
rule-of-thirds alignment. A user study with photographers
and visual artists is the natural next step, both to
validate these metrics against human judgment and to
evaluate anchor conditioning as a creative interface in
practice. Beyond user evaluation, we plan to extend the
framework in two directions: a complementary anchor
component based on object detection, where a bounding-box
centroid encoded through the same Fourier pathway would
condition the model on subject placement and enable
compositional control at both the scene and subject
levels; and high-resolution generation, where preserving
the anchor across upscaling stages requires a refinement
model trained to treat the anchor as a geometric
constraint, with a wavelet-domain objective separating
fidelity loss on low-frequency subbands from texture loss
on high-frequency subbands \cite{korkmaz2024}.

\textbf{Limitations.} The anchor vector is designed for
landscape imagery where composition is governed primarily
by the horizon and the sky-to-ground relationship.
Portrait, architectural, and product photography are
shaped by different spatial conventions and would require
category-specific anchor definitions to benefit from the
same framework. The automatic metrics used here also do
not capture the full subjective experience of a creative
practitioner working with the system, which is why the
planned user evaluation is central to future iterations
of this work.

Taken together, these results show that a four dimensional
compositional descriptor can condition spatial layout when
given a dedicated pathway, and that category specific
training improves precision over mixed training
\cite{govind2024genai}.






\bibliographystyle{iccc}
\bibliography{iccc}


\end{document}